\documentclass{article}
\usepackage[a4paper, margin=1.0in]{geometry}
\usepackage{url}
\usepackage{amsmath, float, booktabs, multirow, graphicx, placeins, amssymb}
\usepackage{booktabs, multirow, graphicx, amsmath}
\usepackage{adjustbox}
\usepackage{caption}
\title{Tangma: A Tanh-Guided Activation Function with Learnable Parameters}
\author{
   Shreel Golwala\footnote{\protect\url{https://github.com/shreelg/Tangma}} \\
  Virginia Tech \\
  \texttt{golwalas@vt.edu} \\
  \vspace{0.3em}
}
\date{}

\begin{document}
\maketitle
\vspace{-7pt}

%abstract
\section{Abstract}

Activation functions are key to effective backpropagation and expressiveness in deep neural networks. This work introduces Tangma, a new activation function that combines the smooth shape of the hyperbolic tangent with two learnable parameters --- $\alpha$, which shifts the curve’s inflection point to adjust neuron activation, and $\gamma$, which adds linearity to preserve weak gradients and improve training stability. Tangma was evaluated on MNIST and CIFAR-10 using custom networks composed of convolutional and linear layers and compared against ReLU, Swish, and GELU. On MNIST, Tangma achieved the highest validation accuracy of 99.09\% and the lowest validation loss, demonstrating faster and more stable convergence than the baselines. In CIFAR-10, Tangma reached a top validation accuracy of 78.15\%, outperforming all other activation functions while maintaining a competitive training loss. Furthermore, Tangma showed improved training efficiency with lower average epoch runtimes compared to Swish and GELU. These results show that Tangma performs well on standard vision tasks and offers reliable, efficient training. Its learnable design gives more control over activation behavior, which may help larger models learn more consistently in tasks such as image recognition or language modeling.

\section{Introduction}
\vspace{2pt}

\noindent
Activation functions are fundamental to the success of modern deep learning architectures. By introducing non-linearity into the network, they help neural models approximate complex functions beyond the capabilities of linear transformations. 

\vspace{4pt}

\noindent
In a standard neural network, each neuron computes an affine transformation:
\[
z = w^\top x + b,
\]
where \( w \in \mathbb{R}^n \) is the weight vector, \( x \in \mathbb{R}^n \) is the input (either raw data or the output from a previous layer), and \( b \in \mathbb{R} \) is the bias term.

\vspace{4pt}

\noindent
To introduce non-linearity, the output of this affine transformation is passed through a non-linear activation function \( \theta : \mathbb{R} \to \mathbb{R} \), resulting in:
\[
a = \theta(z) = \theta(w^\top x + b).
\]
This non-linearity is applied element-wise across the outputs of an entire layer. For a layer with multiple neurons, the operation generalizes to:
\[
a = \theta(Wx + b),
\]
where \( W \in \mathbb{R}^{m \times n} \) is the weight matrix and \( b \in \mathbb{R}^m \) is the bias vector.

\vspace{4pt}

\noindent
The activation function \( \theta \) is essential for enabling neural networks to learn non-trivial input-output mappings; without it, a stacked multilayer network collapses into a equivalent single-layer linear model due to compositional closure of linear operations, e.g.,
\[
\theta_{\text{identity}}(W_3(W_2(W_1x + b_1) + b_2) + b_3) = W'x + b',
\]
rendering it incapable of learning non-linear patterns or forming complex decision boundaries --- limiting its use for real-world tasks.

\vspace{4pt}

\noindent
In addition to providing expressiveness, activation functions significantly influence training by affecting the flow of gradients during backpropagation. During training, gradients are propagated layer by layer using the chain rule. For a single neuron, the gradient of the loss \( \mathcal{L} \) with respect to the weights is:
\[
\frac{\partial \mathcal{L}}{\partial w} = \frac{\partial \mathcal{L}}{\partial a} \cdot \theta'(z) \cdot x,
\]
where \( \theta'(z) \) is the activation function's derivative at the input \(z\). 

\vspace{4pt}

\noindent
If \( \theta'(z) \) is very small, the gradients vanish and learning stops; if too large, gradients explode, destabilizing training.

% In a standard neural network, each neuron computes an affine transformation \( z = w^\top x + b \), where \( w \in \mathbb{R}^n \) is the weight vector, \( x \in \mathbb{R}^n \) is the input (either raw data or the output from a previous layer), and \( b \in \mathbb{R} \) is the bias term. 

% To introduce non-linearity, the output of the affine transformation passed through a nonlinear activation function \( \theta \)

% The activation function \(\theta(z) \), where \( \theta \) is a non-linear transformation, introduces non-linearity and modulates the neuron output, allowing the network to capture non-trivial input-output mappings. Furthermore, activation functions significantly influence the training dynamics by affecting the flow of gradients during backpropagation.

\vspace{7pt}

\subsection*{2.1 ReLU}

Among commonly used activations, the Rectified Linear Unit (ReLU)\textsuperscript{[1, 2]} is defined as:
\[
\text{ReLU}(x) = \max(0, x),
\]
with derivative:
\[
\frac{d}{dx} \text{ReLU}(x) = 
\begin{cases}
1 & \text{if } x > 0, \\
0 & \text{otherwise}.
\end{cases}
\]
ReLU is computationally efficient and widely adopted due to its ability to mitigate vanishing gradients in the positive domain. However, it suffers from the "dying neuron" problem, wherein neurons exposed to persistently negative inputs output zero and receive zero gradients, resulting in stagnant weights and limited learning in affected regions.

\vspace{7pt}

\subsection*{2.2 GELU}

The Gaussian Error Linear Unit (GELU)\textsuperscript{[4]} is a smooth and probabilistic alternative defined as:
\[
\text{GELU}(x) = x \cdot \Phi(x) = \frac{x}{2} \left(1 + \text{erf}\left(\frac{x}{\sqrt{2}}\right)\right),
\]
where \( \Phi(x) \) denotes the cumulative distribution function of the standard normal distribution. GELU softly gates inputs based on their magnitude and sign, making it effective in transformer-based models. However, its reliance on the Gaussian error function and exponential computation increases its complexity and training time. Furthermore, for moderate negative input (e.g. \( x \in [-3, -1] \)), GELU yields small gradients, leading to slower convergence due to attenuated learning signals. The probabilistic formula also hinders interpretability compared to more deterministic activations.

\vspace{7pt}

\subsection*{2.3 Swish}
Swish, another modern activation function, is defined as:
\[
\text{Swish}(x) = x \cdot \sigma(x) = \frac{x}{1 + e^{-x}},
\]
where \( \sigma(x) \) is the sigmoid function. Swish\textsuperscript{[5]} is smooth and non-monotonic, offering empirical improvements in deep networks by allowing gradients to pass through small negative inputs.  However, for large negative \( x \), since \( \sigma(x) \to 0 \), the gradient \( \frac{d}{dx} \text{Swish}(x) \) also approaches zero, which limits the magnitude of the update.

\vspace{7pt}

\subsection*{2.4 Overview of Tangma}

To address these limitations, this work proposes Tangma, which is defined as: \[\text{Tangma}(x) = x \cdot \tanh(x + \alpha) + \gamma x,\] a novel activation function that combines smooth saturation of the hyperbolic tangent with tunable linearity through learnable parameters\textsuperscript{[6, 7]}. Tangma is designed to improve gradient propagation in deep networks and support a more stable convergence by controlling when and how neurons respond during training.

\section{Mathematical Analysis }

The Tangma activation function is defined as:
\[
\text{Tangma}(x) = x \cdot \tanh(x + \alpha) + \gamma x,
\]
where \( \alpha \in \mathbb{R} \) is a learnable non-linearity shift and \( \gamma \in \mathbb{R} \) is a learnable linear skip coefficient. This function is smooth and differentiable everywhere, enabling stable gradient flow and flexible non-linear modeling across different input ranges.

\subsection*{3.1 Derivative}

The derivative of Tangma with respect to \( x \) is:
\[
\frac{d}{dx} \text{Tangma}(x) = \tanh(x + \alpha) + x \cdot \text{sech}^2(x + \alpha) + \gamma,
\]
where \( \text{sech}(x) = \frac{1}{\cosh(x)} \). This derivative is non-zero as long as \( \gamma \neq 0 \), guaranteeing that the neuron remains active and gradients continue to flow.

\begin{figure}[H]
    \centering
    \includegraphics[width=0.9\linewidth]{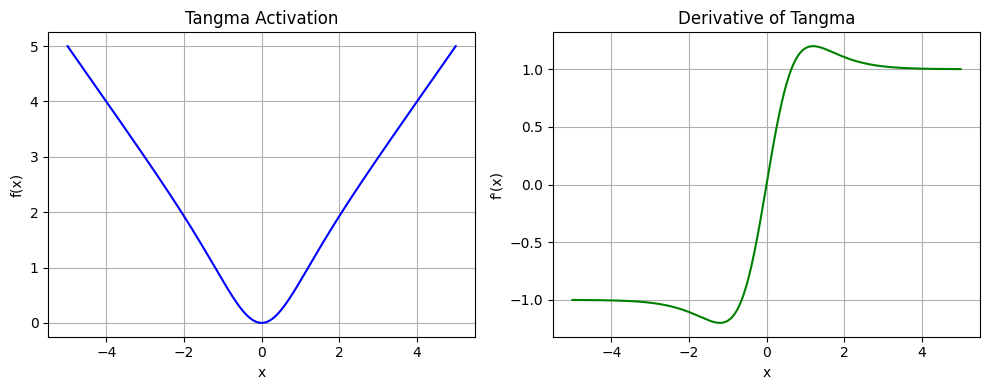}
    \caption{Plot of the Tangma activation function and its derivative for $\alpha = 0$ and $\gamma = 0$.}
    \label{fig:function-derivative-graph}
\end{figure}

\subsection*{3.2 Role of \(\gamma\)}
The coefficient \( \gamma \) ensures a consistent linear gradient path, even when the nonlinear component saturates. This acts like a skip connection, helping gradients propagate smoothly and preserving trainability across all input values. The behavior of Tangma is analyzed in three input conditions:

\vspace{5pt}
\begin{description}

\item[$\bullet$ Small Inputs (\( x \approx 0 \)):] 

Around zero, the function can be approximated by \( \tanh(x + \alpha) \) using a first-order Taylor expansion:
\[
\tanh(x + \alpha) \approx \tanh(\alpha) + x \cdot \text{sech}^2(\alpha),
\]
where \( \text{sech}^2(\alpha) = 1 - \tanh^2(\alpha) \) measures the local slope of the tanh function at \( \alpha \). This expansion expresses the output as a linear function in \( x \) plus higher-order terms.

Substituting this into Tangma,
  \[
    \text{Tangma}(x) = x \cdot \big(\tanh(x + \alpha) + \gamma\big) \approx x \cdot \big(\tanh(\alpha) + \gamma\big) + x^2 \cdot \text{sech}^2(\alpha).
  \]
  Since \( x \) is small, the quadratic term \( x^2 \cdot \text{sech}^2(\alpha) \) is negligible, so the function behaves approximately as
  \[
    \text{Tangma}(x) \approx x \cdot \big(\tanh(\alpha) + \gamma\big).
  \]
  This linearity near zero avoids flat regions in the activation, ensuring neurons remain responsive and gradients do not vanish early in training.

  \item[$\bullet$ Large Negative Inputs (\( x \to -\infty \)):] 

  For very large negative inputs, the argument \( x + \alpha \) tends to \(-\infty\), causing the hyperbolic tangent to saturate at its lower bound:
  \[
    \tanh(x + \alpha) \to -1.
  \]
  Therefore, Tangma behaves as
  \[
    \text{Tangma}(x) \to x \cdot (-1 + \gamma) = (\gamma - 1) x.
  \]
  This linear asymptote guarantees a nonzero gradient as long as \( \gamma \neq 1 \), preventing the gradient starvation issue.

  \item[$\bullet$ Large Positive Inputs (\( x \to \infty \)):] 

  Similarly, when \( x + \alpha \to \infty \), the tanh saturates at its upper bound:
  \[
    \tanh(x + \alpha) \to 1,
  \]
  and Tangma approximates to
  \[
    \text{Tangma}(x) \to x \cdot (1 + \gamma) = (\gamma + 1) x.
  \]
  This linear growth at large positive inputs keeps gradient magnitudes controlled, preventing exploding gradients and keep training stability.

\end{description}

\begin{figure}[H]
    \centering
    \includegraphics[width=0.55\linewidth]{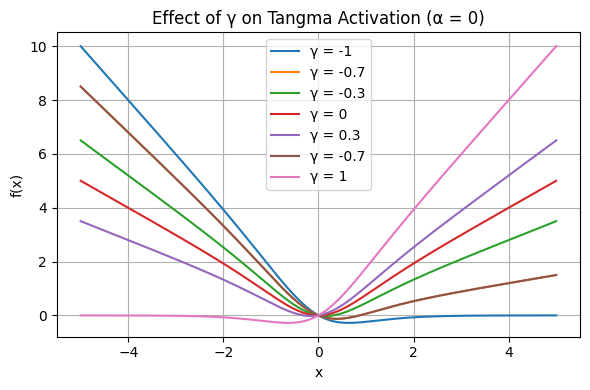}
    \caption{Effect of $\gamma$ on Tangma activation function with $\alpha = 0$.}
    \label{fig:gamma-shift}
\end{figure}

\vspace{10pt}
\subsection*{3.3 Role of \(\alpha\)}
The parameter \( \alpha \) horizontally shifts the nonlinear response of the function. For standard \( \tanh(x) \), the inflection point is at \( x = 0 \); for Tangma, the inflection occurs at \( x = -\alpha \). This shift determines where the neuron transitions between linear behavior and saturation.

\begin{itemize}
    \item When \( \alpha > 0 \), the curve shifts left, activating or saturating earlier.
    \item When \( \alpha < 0 \), the curve shifts right, delaying activation or saturation.
\end{itemize}

\noindent
By learning \( \alpha \), the network can adapt the non-linearity zone to input distribution, biasing neurons toward desired ranges. This increases representational flexibility and allows neurons to adjust where they are most active.

\vspace{10pt}

\begin{figure}[H]
    \centering
    \includegraphics[width=0.55\linewidth]{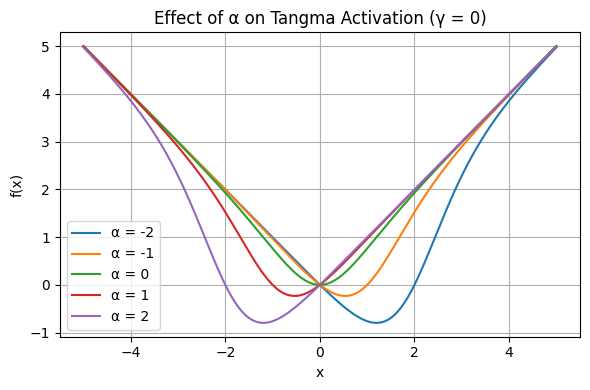}
    \caption{Effect of $\alpha$ on Tangma activation function with $\gamma = 0$.}
    \label{fig:alpha-shift}
\end{figure}

\section{Methodology}

A convolutional neural network was implemented for both MNIST and CIFAR-10, differing only in model depth and input dimensionality. MNIST, being simpler, was trained with a smaller batch size, while the more complex CIFAR-10 used larger batches for convergence. Each model was designed to support four pluggable activation functions --- Swish, GELU, ReLU, and Tangma with $\alpha$ and $\gamma$ set as trainable tensors (\texttt{nn.Parameter}) initialized to \(0\). Each activation function was tested independently on both datasets by training a new model from scratch.

\vspace{4pt}

\noindent
All models were trained using the Adam optimizer with a learning rate of 0.001. The loss function was \texttt{CrossEntropyLoss}. During training, gradients were computed with \texttt{loss.backward()}, and parameters were updated using \texttt{optimizer.step()}, following a reset of previous gradients via \texttt{\detokenize{optimizer.zero_grad()}} to prevent accumulation. Metrics including training loss, validation loss, accuracy, and epoch runtime were recorded for comparison. Training was conducted on a system equipped with an NVIDIA RTX 4060 Mobile GPU using CUDA acceleration.

\vspace{4pt}

\subsection*{4.1 Loss Function and Prediction Rule}
During training, the network outputs a vector of logits, denoted as \[
\mathbf{z} = (z_0, z_1, \ldots, z_{C-1}) \in \mathbb{R}^C,
\] where \textit{C} is the number of classes. For MNIST and CIFAR-10, \textit{C}= 10 as both datasets have 10 categories to classify from.

\vspace{4pt}
\noindent
Given a true class label \( y \in \{0, \ldots, C - 1\} \), the model’s performance is evaluated using the cross-entropy loss, computed directly from the logits using the \textbf{log-sum-exp trick}\textsuperscript{[15]}:

\[
L(\mathbf{}{\textbf{z}}, y) = -z_y + \log\left( \sum_{j=0}^{C-1} e^{z_j} \right).
\]

\noindent
While mathematically equivalent to applying softmax followed by the negative log-likelihood, this approach skips the softmax step, offering the same result with better numerical stability and efficiency, especially when logits are large or vary greatly.

\vspace{4pt}
\noindent
During backpropagation, the loss gradient drives the model to increase the logit corresponding to the correct class \( z_y \), while reducing the logits for incorrect classes.

\vspace{4pt}
\noindent
During inference, the model outputs a logit vector \( \mathbf{z} \), and the predicted class \( \hat{y} \) is the index of the highest logit:

\[
\hat{y} = \arg\max_j z_j.
\]

\vspace{4pt}
\noindent
Since the softmax function is strictly increasing and preserves the order of logits, selecting the highest logit yields the same result as choosing the highest probability. Thus, instead of explicitly computing probabilities that sum to 1, the model can directly use logits for prediction without altering their relative ranking:

\[
z_i > z_j \iff p_i > p_j,
\quad \text{where } p_j = \frac{e^{z_j}}{\sum_{k=0}^{C-1} e^{z_k}}.
\]

\vspace{1pt}

\subsection*{4.2 MNIST architecture}

For MNIST, the data was loaded from Kaggle's MNIST dataset as a CSV file, converted to tensors, and wrapped in a TensorDataset. The input images were grayscale and reshaped from CSV format into 1 × 28 × 28 tensors, with pixel intensities normalized from the range [0, 255] to [0, 1] by dividing all values by 255.  The input tensor 
\[
X \in \mathbb{R}^{64 \times 1 \times 28 \times 28},
\]
representing a batch of 64 grayscale images with spatial dimensions \(28 \times 28\), is first processed by the initial convolutional layer:
\[
\text{Conv2d}_1 : \mathbb{R}^{1 \times 28 \times 28} \rightarrow \mathbb{R}^{32 \times 26 \times 26}.
\]
This layer applies 32 learnable convolutional filters with a kernel size of \(3 \times 3\), stride 1, and no zero-padding, resulting in a reduction of spatial dimensions according to the formula:
\[
\frac{W - K + 2P}{S} + 1 = \frac{28 - 3 + 0}{1} + 1 = 26,
\]
where \(W\) is the input width, \(K\) the kernel size, \(P\) the padding, and \(S\) the stride. The output tensor shape after this operation is:
\[
\mathbb{R}^{64 \times 32 \times 26 \times 26}.
\]
One of the four non-linear activation functions is then applied. 
The subsequent convolutional layer,
\[
\text{Conv2d}_2 : \mathbb{R}^{32 \times 26 \times 26} \rightarrow \mathbb{R}^{64 \times 24 \times 24},
\]
employs 64 filters with the same kernel size and stride, further extracting features while reducing the spatial dimensions by 2 pixels per side. This yields an output tensor of shape \(\mathbb{R}^{64 \times 64 \times 24 \times 24}\), which is again passed through an activation function.

\vspace{4pt}

\noindent
To introduce spatial invariance and reduce computational complexity, a two-dimensional max pooling operation, \texttt{MaxPool2d} with kernel size 2 and stride 2, is applied, halving the spatial resolution to \(12 \times 12\).

\vspace{4pt}

\noindent
The resulting tensor has shape:
\[
\mathbb{R}^{64 \times 64 \times 12 \times 12}.
\]

\noindent
Regularization is applied through a dropout layer with probability \(p = 0.25\), which randomly zeros 25\% of the activations during training to reduce overfitting by preventing co-adaptation of neurons.

\vspace{4pt}
\noindent
The tensor is then flattened into vectors:
\[
z \in \mathbb{R}^{64 \times 9216},
\]

\noindent
where \(9216 = 64 \times 12 \times 12\) corresponds to the total feature dimensions per image.
This vector is passed through a fully connected linear transformation,
\[
\text{FC}_1 : \mathbb{R}^{9216} \rightarrow \mathbb{R}^{128},
\]
which embeds the input into a 128-dimensional latent feature space \(Z \subseteq \mathbb{R}^{128}\). Each neuron in this layer encodes the learned hierarchical features.

\vspace{4pt}
\noindent
The latent representation undergoes a second but the same activation, followed by a dropout layer with an increased dropout probability \(p = 0.5\), to further prevent memorization, as it is in a denser neuron network.

\vspace{4pt}
\noindent
Finally, a second fully connected layer,
\[
\text{FC}_2 : \mathbb{R}^{128} \rightarrow \mathbb{R}^{10},
\]
maps the latent features to logits corresponding to the ten MNIST digit classes. During training, these logits are directly inputted into the loss function, which
computes the classification loss in a numerically stable manner. During inference, the predicted class is chosen as the one with the highest logit score.

\begin{figure}[H]
    \centering
    \includegraphics[width=0.4\linewidth]{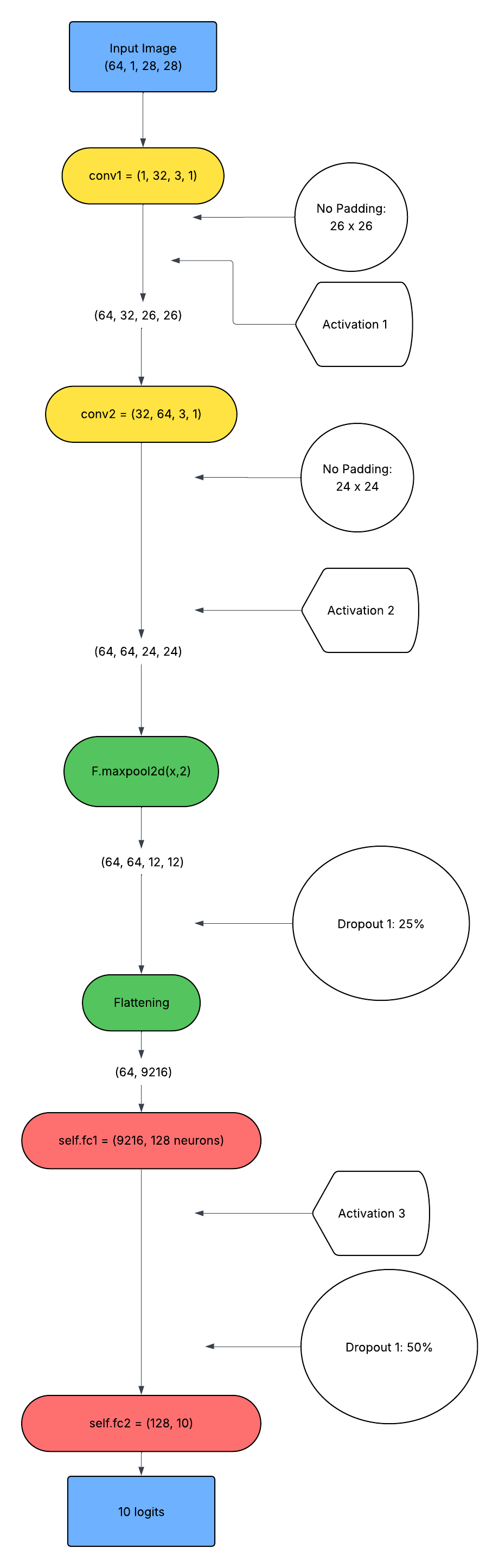}
    \caption{MNIST forward-pass structure.}
    \label{fig:mnist-pass-shift}
\end{figure}

\vspace{7pt}

\subsection*{4.3 CIFAR-10 architecture}

For CIFAR-10, the dataset was loaded using \texttt{torchvision.datasets.CIFAR10} with standard PyTorch \texttt{DataLoader} pipelines.
\[
X \in \mathbb{R}^{128 \times 3 \times 32 \times 32},
\]
representing a batch of 128 RGB images with spatial resolution \(32 \times 32\), is first scaled from the original pixel intensity range \([0, 255]\) to \([0, 1]\) by dividing each value by 255. Then, normalization is applied per channel to map the values to the range \([-1, 1]\) using the transformation:

\[
\hat{x} = \frac{x - \mu}{\sigma} = \frac{[0, 1] - 0.5}{0.5} = [-1, 1],
\]

\vspace{4pt}
\noindent
where \(x\) is the scaled pixel value, \(\mu = 0.5\) is the mean, and \(\sigma = 0.5\) is the standard deviation. This process shifts and scales pixel intensities such that 0 maps to -1, 0.5 maps to 0, and 1 maps to 1.

\vspace{4pt}
\noindent
The initial convolutional layer,
\[
\text{Conv2d}_1 : \mathbb{R}^{3 \times 32 \times 32} \rightarrow \mathbb{R}^{32 \times 32 \times 32},
\]
applies 32 learnable filters of kernel size \(3 \times 3\), stride 1, and padding 1, to preserve the spatial dimensions. This is followed by an element-wise activation and a max pooling operation with kernel size 2 and stride 2, reducing the resolution:
\[
\mathbb{R}^{32 \times 32 \times 32} \rightarrow \mathbb{R}^{32 \times 16 \times 16}.
\]
The second convolutional layer,
\[
\text{Conv2d}_2 : \mathbb{R}^{32 \times 16 \times 16} \rightarrow \mathbb{R}^{64 \times 16 \times 16},
\]
also uses 64 filters with identical kernel size and padding, maintaining spatial size before another activation and max pooling step:
\[
\mathbb{R}^{64 \times 16 \times 16} \rightarrow \mathbb{R}^{64 \times 8 \times 8}.
\]
The third convolutional layer,
\[
\text{Conv2d}_3 : \mathbb{R}^{64 \times 8 \times 8} \rightarrow \mathbb{R}^{128 \times 8 \times 8},
\]
follows the same pattern, and subsequent pooling yields:
\[
\mathbb{R}^{128 \times 8 \times 8} \rightarrow \mathbb{R}^{128 \times 4 \times 4}.
\]
The output is flattened into vectors 
\[
z \in \mathbb{R}^{128 \times 2048},
\]
with \(2048 = 128 \times 4 \times 4\), and passed through a fully connected layer:
\[
\text{FC}_1 : \mathbb{R}^{2048} \rightarrow \mathbb{R}^{512},
\]
followed by an activation and dropout with probability \(p = 0.5\) to prevent overfitting in the dense representation. The final layer performs the classification:
\[
\text{FC}_2 : \mathbb{R}^{512} \rightarrow \mathbb{R}^{10},
\]
producing raw logits. The predicted class is selected in the same manner as for MNIST.

\begin{figure}[H]
    \centering
    \includegraphics[width=0.34\linewidth]{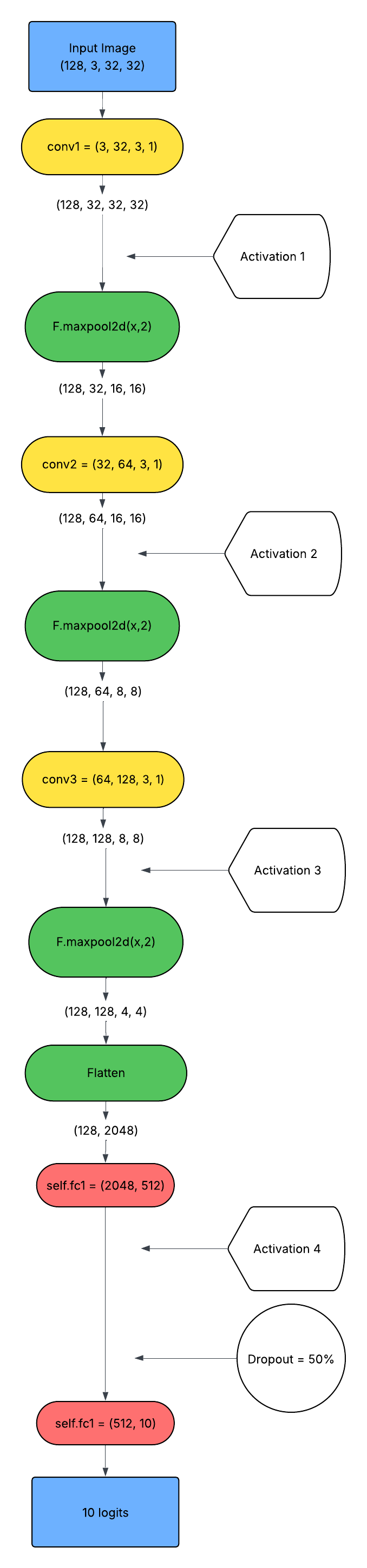}
    \caption{CIFAR-10 forward-pass structure.}
    \label{fig:cifar-forward-shift}
\end{figure}

\vspace{1pt}

\section{Evaluation}

\subsection*{5.1 Performance on MNIST}
Each activation function was used to train a separate but similar model, and results were evaluated on a 12,000-image validation set (from an 80/20 split of the 60,000-image MNIST dataset).

\vspace{22pt}

\begin{figure}[H]
    \centering
    \includegraphics[width=0.6\linewidth]{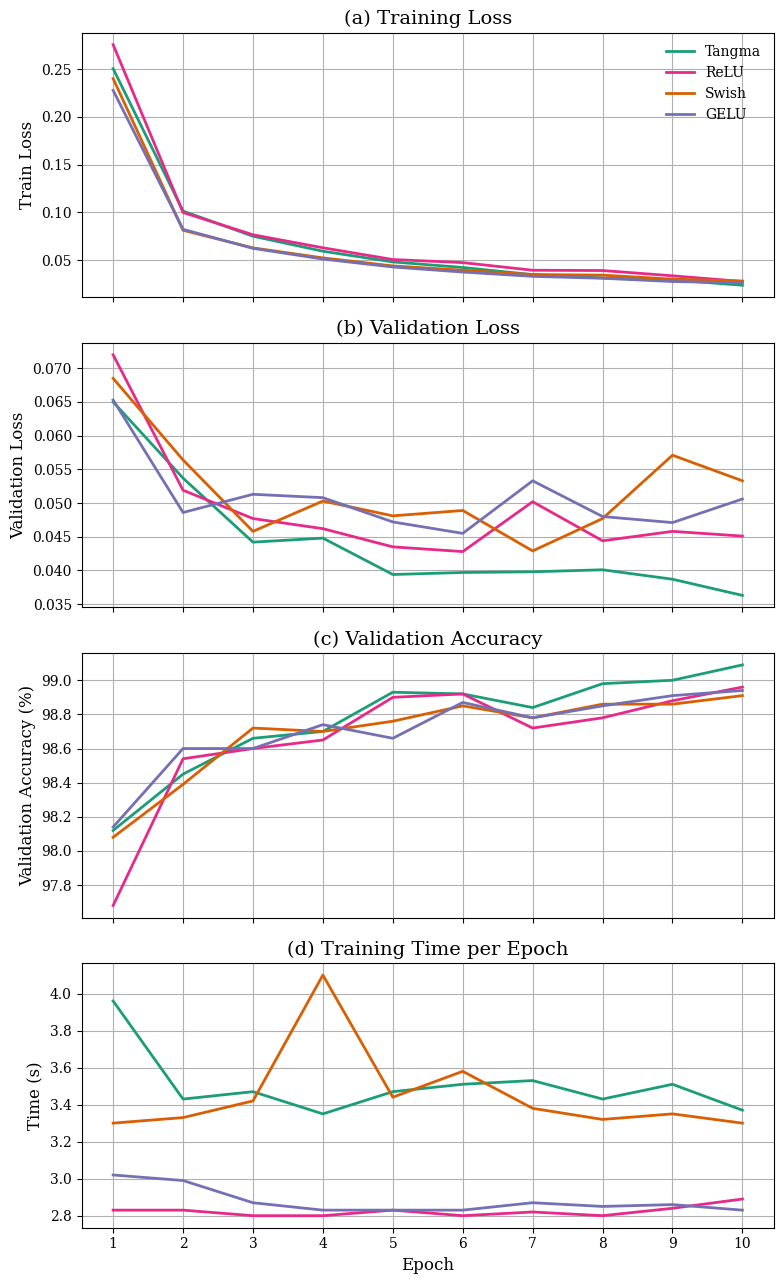}
    \caption{Performance of the four activation functions on MNIST. The plots show training loss (a), validation loss (b), validation accuracy (c), and training time per epoch (d) over 10 epochs.}
    \label{fig:mnist-performance}
\end{figure}

\begin{table}[h]
\centering
\begin{tabular}{lcccc}
\toprule
Activation & Final Val Accuracy (\%) & Final Val Loss & Final Train Loss & Avg Epoch Time (s) \\
\midrule
Tangma & 99.09 & 0.0363 & 0.0239 & 3.45 \\
ReLU   & 98.96 & 0.0451 & 0.0278 & 2.82 \\
Swish  & 98.91 & 0.0533 & 0.0283 & 3.44 \\
GELU   & 98.94 & 0.0506 & 0.0264 & 2.89 \\
\bottomrule
\end{tabular}
\caption{Comparison of final performance metrics of four activation functions on MNIST.}
\label{tab:activation_results}
\end{table}

\subsubsection{Results (see Appendix for full data)}
Tangma shows a consistent, though modest, improvement in final validation accuracy and loss compared to ReLU, Swish, and GELU. It achieves a final validation accuracy of 99.09\%, passing ReLU by 0.13 percentage points, Swish by 0.18, and GELU by 0.15. It also yields the lowest final validation loss (0.0363) and training loss (0.0239), indicating better fit and generalization on the MNIST dataset.

\vspace{4pt}
\noindent
The convergence behavior further highlights Tangma's advantage. By Epoch 3, Tangma reaches a validation accuracy of 98.66\%, slightly outperforming Swish (98.72\%), ReLU (98.60\%), and GELU (98.60\%). From Epoch 3 onward, Tangma maintains a steady lead, achieving faster reduction in training loss, dropping from 0.2505 at epoch 1 to 0.0483 by epoch 5. This suggests improved gradient flow and optimization dynamics that support faster initial convergence. 

\vspace{4pt}
\noindent
After epoch 5, Tangma's validation loss fluctuates within a narrow margin of 0.0038, lower and less volatile compared to the variation observed in other functions --- indicating a better generalization and greater resistance to overfitting. This smoother loss curve aligns with more stable model behavior during training. This smoother loss curve reflects more stable training behavior and improved model consistency. 

\vspace{4pt}
\noindent
In terms of computational efficiency, Tangma's average epoch time is approximately 3.45 seconds --- comparable to Swish (3.44s), though moderately slower than ReLU (2.82s) and GELU (2.89s). Given Tangma's gains in accuracy and training stability, this slight increase in computation time represents a reasonable trade-off for MNIST-scale models. 

\vspace{4pt}
\noindent

\subsubsection{Discussion}
From a Tangma--MNIST interaction standpoint, Tangma shows strong performance by capturing fine-grained, local pixel-level variations important to handwritten digit recognition. Unlike natural image datasets where texture and color dominate, MNIST depends on precise stroke geometry within narrow spatial regions. A further challenge is that MNIST pixel intensities are normalized from $[0, 255]$ to $[0, 1]$, which stabilizes training but compresses structural features into low-magnitude values. 

\vspace{4pt}
\noindent
In early convolutional layers, small or slightly negative activations often arise near thin strokes, junctions, or loop closures---regions that carry subtle yet critical information. Digits often differ through fine-grained features---such as the open or closed top of a ``4,'' the curvature in a ``6,'' or the loop closure in a ``9.'' These involve small shifts in localized intensity without changing the label, requiring sensitivity to fine-scale deformation. This combination is particularly effective for digits with similar structures. Discriminating a ``3'' from an ``8'' relies on detecting whether arcs connect, while telling a narrow ``8'' from an open ``0,'' or avoiding excessive response to a thick, bold stroke in a ``3,'' may depend on loop closure and spacing.

\vspace{4pt}
\noindent
Traditional activations like ReLU discard these subtleties by zeroing out low signals, potentially losing cues needed to distinguish digits. Tangma addresses this by blending a smooth $\tanh(x + \alpha)$ term, which saturates overly strong responses, with a linear $\gamma x$ term that retains low-magnitude activations, preserving subtle features.

\subsection*{5.2 Performance on CIFAR-10}
Similar to MNIST, models for CIFAR-10 were assessed on a validation set derived from a 90/10 split of the 60,000-image dataset, with 6,000 images reserved for validation.

\begin{figure}[H]
    \centering
    \includegraphics[width=0.6\linewidth]{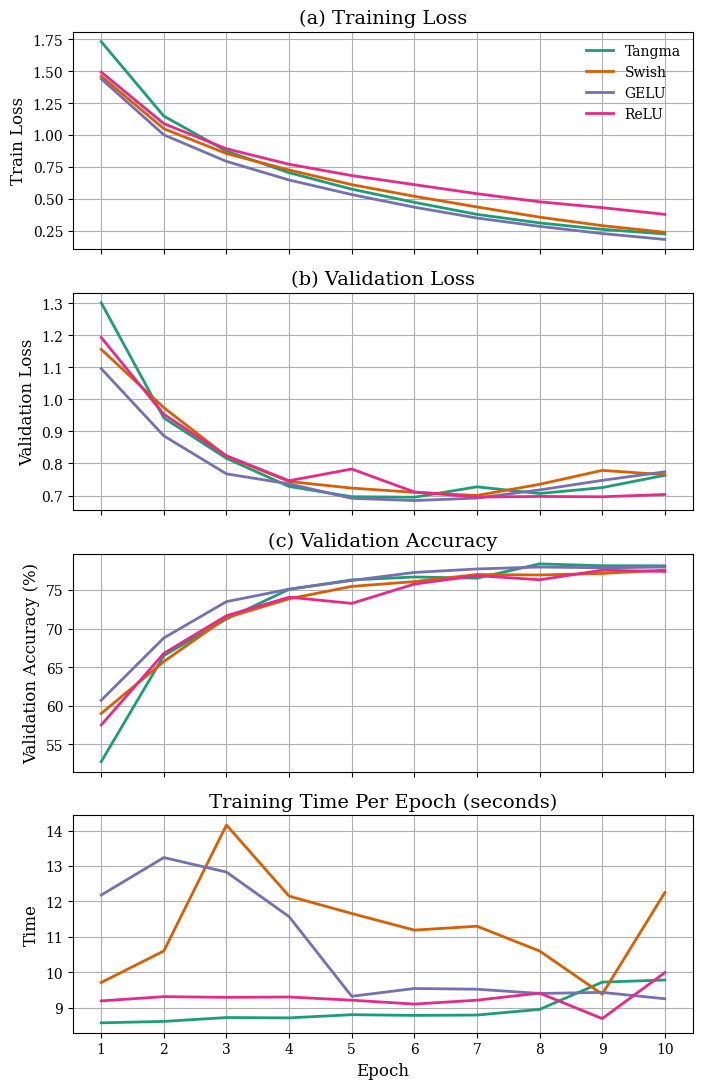}
    \caption{Performance of the four activation functions on CIFAR-10. The plots show training loss (a), validation loss (b), validation accuracy (c), and training time per epoch (d) over 10 epochs.}
    \label{fig:cifar-performance}
\end{figure}

\begin{table}[h]
\centering
\begin{tabular}{lcccc}
\toprule
Activation & Final Val Accuracy (\%) & Final Val Loss & Final Train Loss & Avg Epoch Time (s) \\
\midrule
Tangma & 78.15 & 0.7631 & 0.2270 & 8.97 \\
Swish  & 77.59 & 0.7649 & 0.2400 & 11.20 \\
GELU   & 77.99 & 0.7740 & 0.1840 & 11.32 \\
ReLU   & 77.42 & 0.7032 & 0.3802 & 9.35 \\
\bottomrule
\end{tabular}
\caption{Comparison of final performance metrics of four activation functions on CIFAR-10.}
\label{tab:cifar_results}
\end{table}

\subsubsection{Results (see Appendix for full data)}

Tangma also displays a balance of accuracy, convergence speed, and computational efficiency on CIFAR-10, though its advantages are more modest here than on MNIST. by Epoch 5, Tangma already reaches 76.32\% validation accuracy, outpacing  ReLU (73.28\%), Swish (75.48\%), and even narrowly beating GELU (76.25\%). During the 10 epochs, Tangma reaches the highest final accuracy of 78.15\%, compared to 77.42\% for ReLU, 77.59\% for Swish, and 77.99\% for GELU, a gain of 0.16 to 0.73 percentage points. Its training loss drops steeply from 1.7313 in Epoch 1 to 0.2270 by Epoch 10, undercutting or matching other activations (GELU reaches 0.1840, Swish 0.2400, ReLU 0.3802).

\vspace{4pt}
\noindent
However, Tangma’s validation loss plateaus around 0.69–0.76 after Epoch 5 and finishes at 0.7631, slightly above ReLU’s 0.7032 and GELU’s 0.7740, and comparable to Swish’s 0.7649. This suggests that Tangma may incur a small over-confidence penalty as its correct predictions come with weaker probabilities despite having the best raw accuracy.

\vspace{4pt}
\noindent
In terms of efficiency, Tangma averages 8.97 seconds per epoch, faster than Swish (11.2 s) and GELU (11.3 s), and marginally quicker than ReLU (9.4 s). This makes it the most computationally efficient of the four, achieving top accuracy with the lowest runtime.

\subsubsection{Discussion}

The \(\tanh(x + \alpha)\) component of Tangma preserves low-magnitude activations from fine details in early convolutional layers. For example, in birds, faint edges like claws and beaks appear as weak gradients against noisy backgrounds; the \(\tanh\) smoothly activates here, maintaining these subtle boundaries. Similarly, in cats and dogs, small texture and shape differences produce low-intensity pixel variations, which the symmetric \(\tanh(x + \alpha)\) retains across feature maps.

\vspace{4pt}
\noindent
In deer images, antlers and legs blend into leafy environments, creating shallow gradients that \(\tanh\) preserves to maintain spatial continuity and separate objects from backgrounds. Frogs’ dense skin textures generate rapid but low-magnitude activations; these pass through the linear \(\gamma x\) term, which amplifies them without clipping, keeping texture-sensitive filters active.

\vspace{4pt}
\noindent
For airplanes with long, low-contrast outlines against skies, the activation curve keeps weak horizontal edges active even when gradients are small. In contrast, automobiles and trucks produce mid- to high-magnitude activations; here, \(\tanh\) saturates values to prevent spikes and stabilize responses. Sharp divisions, like truck cabs and trailers, are softened by saturation without losing structure.

\vspace{4pt}
\noindent
Some categories combine soft and sharp features, requiring an activation function that handles both extremes. Ships, for instance, have weak edges near horizons alongside strong hull transitions, while dogs mix smooth contours with sharper boundaries such as ears or limbs. Tangma’s \(\gamma x\) and \(x + \alpha\) terms preserve weak signals by modulating activation thresholds, while \(\tanh\) stabilizes strong contrasts, preventing misclassification from exaggerated activations.

\subsection*{5.3 Behavior of Learnable Parameters}

For both models, the learnable parameters $\alpha$ and $\gamma$ were recorded at batches 130 and 260, roughly the midpoint and end of each epoch, during training. This sampling provides reasonable granularity to see how the function changes over time and check its stability within each epoch.

\begin{figure}[H]
    \centering
    \includegraphics[width=0.6\linewidth]{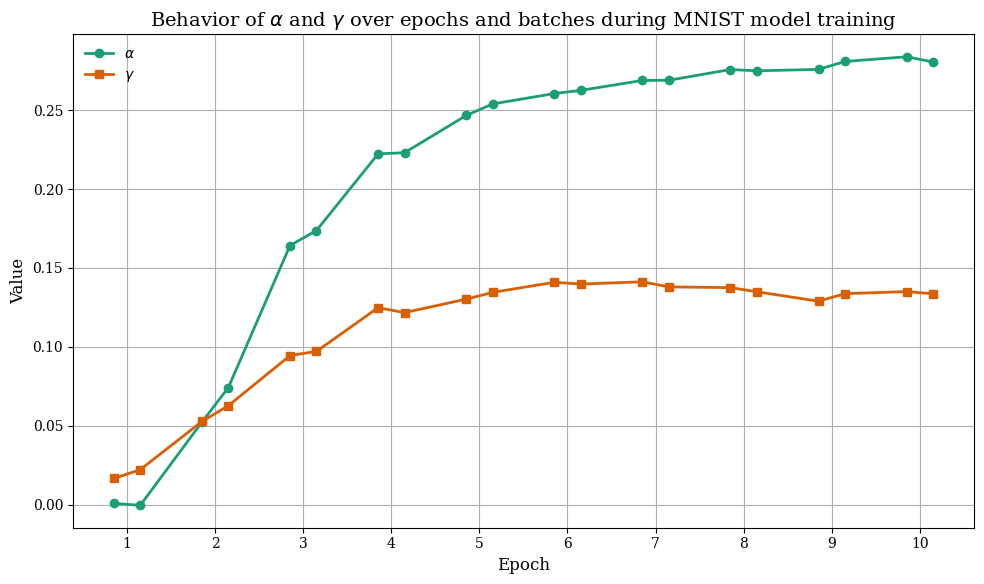}
    \caption{Behavior of parameters $\alpha$ and $\gamma$ on MNIST across 10 epochs.}
    \label{fig:mnist-parameters}
\end{figure}

\begin{figure}[H]
    \centering
    \includegraphics[width=0.6\linewidth]{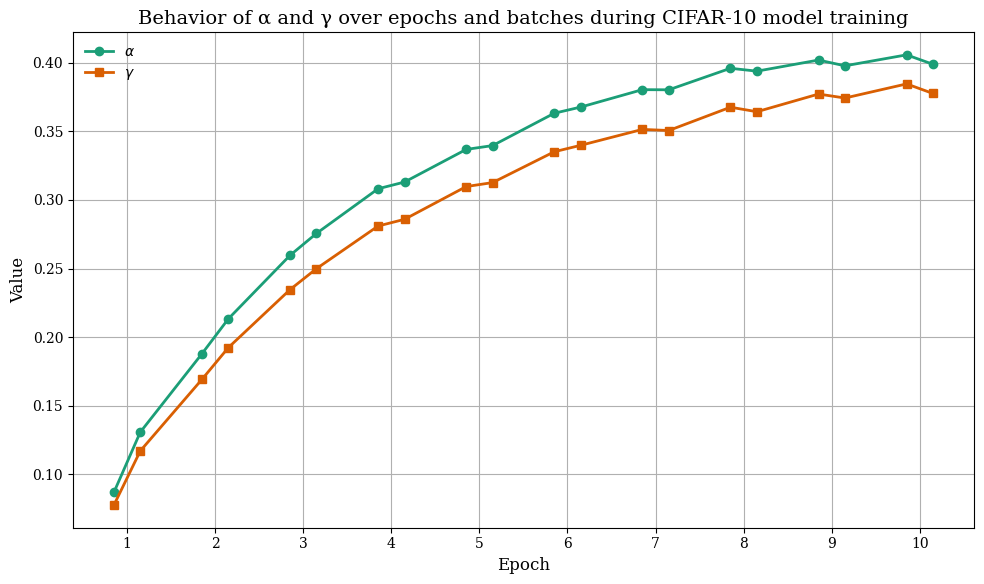}
    \caption{Behavior of parameters $\alpha$ and $\gamma$ on CIFAR-10 across 10 epochs.}
    \label{fig:cifar-parameters}
\end{figure}

\noindent
Throughout the training, the evolution of the learnable parameters reflect how model shifts its focus from fine edge preservation to saturation control, depending on the dataset's visual complexity.

\vspace{4pt}
\noindent
On MNIST, with its low-resolution grayscale images, early layer activations tend to be weak and localized due to the thin strokes, junctions, and simple geometric features. In response, $\gamma$ begins at a low value (0.016) but increases quickly by the second epoch (0.062), intensifying the contribution of low-magnitude activations that are otherwise prone to suppression. During this initial phase, $\alpha$ remains close to zero, since activations rarely reach saturation and most computations operate within the linear range of $\tanh(x + \alpha)$.

\vspace{4pt}

\noindent
Between epochs 3 and 6, both $\alpha$ and $\gamma$ increase steadily --- $\gamma$ reaching 0.141 and $\alpha$ moving from 0.173 to 0.262. This corresponds to the model beginning to recognize more structured patterns, such as loops, curved edges, and stroke combinations that generate moderately strong activations. The rise in $\alpha$ serves to regulate these larger signals by gradually shifting the saturation range, while $\gamma$ helps maintain sensitivity to lower-range features. By the final epochs, both parameters plateau --- $\alpha$ around 0.283 and $\gamma$ near 0.135 --- indicating a stable balance between controlling high activations and preserving detail, as the network becomes consistent in representing full character structure.

\vspace{4pt}

\noindent
In contrast, CIFAR-10 involves higher-resolution RGB images with more textures, color distributions, and frequent object-background overlap. As a result, both parameters start at higher values even in the first epoch ($\alpha = 0.087$, $\gamma = 0.077$), reflecting the need to both suppress noisy, high-magnitude signals and boost mid-level activations. By epoch 3, $\alpha$ and $\gamma$ reach 0.276 and 0.250 --- values that MNIST only reaches near convergence --- indicating the network’s faster adaptation to diverse and cluttered visual inputs.

\vspace{4pt}

\noindent
From epoch 4 onward, parameter growth continues, with $\alpha$ stabilizing near 0.399 and $\gamma$ around 0.377 by epoch 10. This trend reflects the model’s requirement to regulate a broader range of activations, because of overlapping textures, lighting variations, and subtle structural transitions common in natural scenes.

\section{Conclusion}
Tangma demonstrated strong performance across benchmark tests, due to its learnable parameters that balance saturation and sensitivity. This balance preserves gradient flow and feature diversity, helping to mitigate overfitting to stronger signals while retaining weaker but important features. Future work could explore adding Tangma into deeper convolutional and residual architectures as well as evaluating its effectiveness on large-scale vision and language tasks, including transformer-based models.

%resnet or densenet

 % $\gamma$ adjusts the linear sensitivity to subtle input variations, while $\alpha$ controls the saturation threshold to prevent large activations from dominating the representation. This coordinated adjustment allows the activation function to remain responsive across layers while avoiding overfitting to strong features or under-reacting to weaker cues.

\section{References}
\renewcommand{\refname}{}

\section{Appendix}
 
\vspace{-1em}

\setlength{\textfloatsep}{10pt plus 1.0pt minus 2.0pt} % tighten float spacing
\setlength{\floatsep}{6pt plus 1.0pt minus 2.0pt}
\setlength{\intextsep}{8pt plus 1.0pt minus 2.0pt}

\begin{table}[htbp]
\centering
\caption{Training and validation metrics across epochs for different activation functions on MNIST.
TL: Train Loss, VL: Val Loss, VA: Val Accuracy, T: Time.}
\label{tab:mnist-metrics}
\resizebox{\textwidth}{!}{%
\begin{tabular}{c|cccc|cccc|cccc|cccc}
\toprule
\multirow{2}{*}{Epoch} 
& \multicolumn{4}{c|}{Tangma} 
& \multicolumn{4}{c|}{ReLU} 
& \multicolumn{4}{c|}{Swish} 
& \multicolumn{4}{c}{GELU} \\
& TL & VL & VA (\%) & T (s) 
& TL & VL & VA (\%) & T (s) 
& TL & VL & VA (\%) & T (s) 
& TL & VL & VA (\%) & T (s) \\
\midrule
1  & 0.2505 & 0.0650 & 98.12 & 3.96 & 0.2756 & 0.0720 & 97.68 & 2.83 & 0.2401 & 0.0685 & 98.08 & 3.30 & 0.2278 & 0.0653 & 98.14 & 3.02 \\
2  & 0.1015 & 0.0537 & 98.45 & 3.43 & 0.1000 & 0.0519 & 98.54 & 2.83 & 0.0814 & 0.0564 & 98.39 & 3.33 & 0.0824 & 0.0486 & 98.60 & 2.99 \\
3  & 0.0752 & 0.0442 & 98.66 & 3.47 & 0.0767 & 0.0477 & 98.60 & 2.80 & 0.0630 & 0.0458 & 98.72 & 3.42 & 0.0625 & 0.0513 & 98.60 & 2.87 \\
4  & 0.0595 & 0.0448 & 98.70 & 3.35 & 0.0631 & 0.0462 & 98.65 & 2.80 & 0.0525 & 0.0503 & 98.70 & 4.10 & 0.0513 & 0.0508 & 98.74 & 2.83 \\
5  & 0.0483 & 0.0394 & 98.93 & 3.47 & 0.0508 & 0.0435 & 98.90 & 2.83 & 0.0441 & 0.0481 & 98.76 & 3.44 & 0.0430 & 0.0472 & 98.66 & 2.83 \\
6  & 0.0425 & 0.0397 & 98.92 & 3.51 & 0.0476 & 0.0428 & 98.92 & 2.80 & 0.0396 & 0.0489 & 98.85 & 3.58 & 0.0377 & 0.0455 & 98.87 & 2.83 \\
7  & 0.0350 & 0.0398 & 98.84 & 3.53 & 0.0396 & 0.0502 & 98.72 & 2.82 & 0.0351 & 0.0429 & 98.78 & 3.38 & 0.0332 & 0.0533 & 98.78 & 2.87 \\
8  & 0.0316 & 0.0401 & 98.98 & 3.43 & 0.0393 & 0.0444 & 98.78 & 2.80 & 0.0345 & 0.0477 & 98.86 & 3.32 & 0.0312 & 0.0480 & 98.85 & 2.85 \\
9  & 0.0295 & 0.0387 & 99.00 & 3.51 & 0.0338 & 0.0458 & 98.88 & 2.84 & 0.0303 & 0.0571 & 98.86 & 3.35 & 0.0278 & 0.0471 & 98.91 & 2.86 \\
10 & 0.0239 & 0.0363 & 99.09 & 3.37 & 0.0278 & 0.0451 & 98.96 & 2.89 & 0.0283 & 0.0533 & 98.91 & 3.30 & 0.0264 & 0.0506 & 98.94 & 2.83 \\
\bottomrule
\end{tabular}
}
\end{table}

\vspace{-1pt}

\begin{table}[htbp]
\centering
\caption{Training and validation metrics across epochs for different activation functions on CIFAR-10. TL: Train Loss, VL: Val Loss, VA: Val Accuracy, T: Time.}
\label{tab:cifar-metrics}
\resizebox{\textwidth}{!}{%
\begin{tabular}{c|cccc|cccc|cccc|cccc}
\toprule
\multirow{2}{*}{Epoch} 
& \multicolumn{4}{c|}{Tangma} 
& \multicolumn{4}{c|}{Swish} 
& \multicolumn{4}{c|}{GELU} 
& \multicolumn{4}{c}{ReLU} \\
& TL & VL & VA (\%) & T (s) 
& TL & VL & VA (\%) & T (s) 
& TL & VL & VA (\%) & T (s) 
& TL & VL & VA (\%) & T (s) \\
\midrule
1 & 1.7313 & 1.3016 & 52.76 & 8.57 & 1.4615 & 1.1562 & 59.01 & 9.71 & 1.4403 & 1.0960 & 60.72 & 12.18 & 1.4932 & 1.1932 & 57.52 & 9.19 \\
2 & 1.1485 & 0.9421 & 66.51 & 8.61 & 1.0499 & 0.9742 & 65.74 & 10.60 & 1.0019 & 0.8861 & 68.80 & 13.24 & 1.0897 & 0.9533 & 66.80 & 9.31 \\
3 & 0.8750 & 0.8160 & 71.26 & 8.72 & 0.8560 & 0.8226 & 71.43 & 14.16 & 0.7946 & 0.7676 & 73.51 & 12.83 & 0.8931 & 0.8240 & 71.68 & 9.29 \\
4 & 0.7059 & 0.7282 & 75.10 & 8.71 & 0.7270 & 0.7437 & 73.87 & 12.15 & 0.6489 & 0.7363 & 75.12 & 11.57 & 0.7721 & 0.7461 & 74.08 & 9.30 \\
5 & 0.5773 & 0.6962 & 76.32 & 8.80 & 0.6126 & 0.7231 & 75.48 & 11.66 & 0.5351 & 0.6913 & 76.25 & 9.32 & 0.6836 & 0.7827 & 73.28 & 9.21 \\
6 & 0.4742 & 0.6946 & 76.70 & 8.78 & 0.5213 & 0.7099 & 76.10 & 11.19 & 0.4362 & 0.6842 & 77.29 & 9.54 & 0.6126 & 0.7111 & 75.78 & 9.10 \\
7 & 0.3800 & 0.7271 & 76.55 & 8.79 & 0.4381 & 0.7001 & 77.02 & 11.30 & 0.3514 & 0.6920 & 77.74 & 9.52 & 0.5417 & 0.6951 & 76.88 & 9.21 \\
8 & 0.3132 & 0.7062 & 78.40 & 8.95 & 0.3583 & 0.7349 & 76.96 & 10.60 & 0.2863 & 0.7174 & 77.99 & 9.40 & 0.4783 & 0.6971 & 76.34 & 9.41 \\
9 & 0.2628 & 0.7247 & 78.16 & 9.72 & 0.2918 & 0.7784 & 77.16 & 9.38  & 0.2302 & 0.7473 & 77.89 & 9.43 & 0.4326 & 0.6960 & 77.58 & 8.69 \\
10 & 0.2270 & 0.7631 & 78.15 & 9.78 & 0.2400 & 0.7649 & 77.59 & 12.25 & 0.1840 & 0.7740 & 77.99 & 9.25 & 0.3802 & 0.7032 & 77.42 & 9.99 \\
\bottomrule
\end{tabular}
}
\end{table}

\begin{table}[htbp]
\centering
\caption{Learnable Tangma parameters $\alpha$ and $\gamma$ at batches 130 and 260 during MNIST training.}
\label{tab:tangma-params-new}
\begin{tabular}{c|cc|cc}
\toprule
\multirow{2}{*}{Epoch} & \multicolumn{2}{c|}{$\alpha$} & \multicolumn{2}{c}{$\gamma$} \\
 & Batch 130 & Batch 260 & Batch 130 & Batch 260 \\
\midrule
1 & 0.0008 & -0.0004 & 0.0166 & 0.0222 \\
2 & 0.0523 & 0.0742 & 0.0527 & 0.0627 \\
3 & 0.1642 & 0.1737 & 0.0945 & 0.0971 \\
4 & 0.2223 & 0.2230 & 0.1248 & 0.1217 \\
5 & 0.2466 & 0.2539 & 0.1302 & 0.1345 \\
6 & 0.2605 & 0.2625 & 0.1409 & 0.1398 \\
7 & 0.2688 & 0.2689 & 0.1412 & 0.1380 \\
8 & 0.2757 & 0.2749 & 0.1375 & 0.1349 \\
9 & 0.2758 & 0.2808 & 0.1289 & 0.1337 \\
10 & 0.2838 & 0.2805 & 0.1350 & 0.1336 \\
\bottomrule
\end{tabular}
\end{table}

\begin{table}[htbp]
\centering
\caption{Learnable Tangma parameters $\alpha$ and $\gamma$ at batches 130 and 260 during CIFAR-10 training.}
\label{tab:tangma-params}
\begin{tabular}{c|cc|cc}
\toprule
\multirow{2}{*}{Epoch} & \multicolumn{2}{c|}{$\alpha$} & \multicolumn{2}{c}{$\gamma$} \\
 & Batch 130 & Batch 260 & Batch 130 & Batch 260 \\
\midrule
1 & 0.0866 & 0.1306 & 0.0773 & 0.1169 \\
2 & 0.1878 & 0.2133 & 0.1691 & 0.1923 \\
3 & 0.2596 & 0.2756 & 0.2346 & 0.2499 \\
4 & 0.3082 & 0.3131 & 0.2810 & 0.2859 \\
5 & 0.3369 & 0.3397 & 0.3098 & 0.3127 \\
6 & 0.3633 & 0.3678 & 0.3352 & 0.3399 \\
7 & 0.3805 & 0.3804 & 0.3515 & 0.3507 \\
8 & 0.3961 & 0.3940 & 0.3677 & 0.3644 \\
9 & 0.4021 & 0.3979 & 0.3773 & 0.3744 \\
10 & 0.4059 & 0.3990 & 0.3847 & 0.3777 \\
\bottomrule
\end{tabular}
\end{table}

\end{document}